\documentclass{article}
\usepackage{ijcai25}

\usepackage{times}
\usepackage{soul}
\usepackage{url}
\usepackage[hidelinks]{hyperref}
\usepackage[utf8]{inputenc}
\usepackage[small]{caption}
\usepackage{graphicx}
\usepackage{amsmath}
\usepackage{amsthm}
\usepackage{booktabs}
\usepackage{algorithm}
\usepackage{algorithmic}
\usepackage[switch]{lineno}
\usepackage{multirow}
\usepackage{amssymb}
\usepackage{booktabs}
\usepackage{xcolor}
\usepackage{subcaption}
\title{Multimodal Inverse Attention Network with Intrinsic Discriminant Feature Exploitation for Fake News Detection}

\author{
Tianlin Zhang$^1$\thanks{Equal contribution.}\and
En Yu$^2$\footnotemark[1]\and
Yi Shao$^1$\and
Jiande Sun$^1$\thanks{Corresponding author.}
\\
\affiliations
$^1$School of Information Science and Engineering, Shandong Normal University\\
$^2$Faculty of Engineering \& Information Technology, University of Technology Sydney\\
\emails
taylorzhang19951019@gmail.com,
isenn.yu@gmail.com,
yi.shao.mail@gmail.com,
jiandesun@hotmail.com
}
\begin{document}

\maketitle

\begin{abstract}
 Multimodal fake news detection has garnered significant attention due to its profound implications for social security. While existing approaches have contributed to understanding cross-modal consistency, they often fail to leverage modal-specific representations and explicit discrepant features. To address these limitations, we propose a Multimodal Inverse Attention Network (MIAN), a novel framework that explores intrinsic discriminative features based on news content to advance fake news detection. Specifically, MIAN introduces a hierarchical learning module that captures diverse intra-modal relationships through local-to-global and local-to-local interactions, thereby generating enhanced unimodal representations to improve the identification of fake news at the intra-modal level. Additionally, a cross-modal interaction module employs a co-attention mechanism to establish and model dependencies between the refined unimodal representations, facilitating seamless semantic integration across modalities. To explicitly extract inconsistency features, we propose an inverse attention mechanism that effectively highlights the conflicting patterns and semantic deviations introduced by fake news in both intra- and inter-modality. Extensive experiments on benchmark datasets demonstrate that MIAN significantly outperforms state-of-the-art methods, underscoring its pivotal contribution to advancing social security through enhanced multimodal fake news detection.
\end{abstract}

\section{Introduction}
\label{introduction}
The rapid evolution of information technology and social networks has shifted news consumption from traditional media to online platforms, enabling individuals to act as publishers and accelerating the delivery of fast, diverse, and personalized information. However, it has also enabled the spread of misinformation, frequently driven by misinterpretations, exaggerations, or deliberate falsifications, which pose significant risks to social security and stability \cite{shu2017fake}.  Moreover, advancements in GenAI models \cite{ouyang2022training,li2022blip} have further lowered the barriers to creating and distributing fake news, exceeding the capacity of human judgment to identify disinformation effectively \cite{wang2024deepfaker}. While relevant security departments have established verification mechanisms to combat fake news, the vast amount of content has made manual verification increasingly unfeasible \cite{zhou2020survey}. Consequently, the development of automated fake news detection techniques has emerged as a critical research focus to safeguard the reliability and integrity of public information.

\begin{figure}[!t]
    \centering
    \includegraphics[width=\columnwidth]{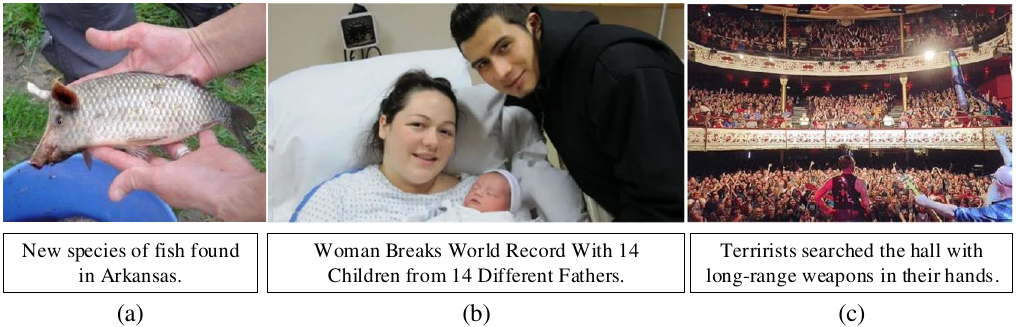}
    \caption{Typical types of fake news. (a)/(b) fabricated fake news where either the text or image is authentic while the other is manipulated to create fake content. Specifically, in (a), the subject in the image features a fish’s body alongside a pig’s ears and snout, highlighting categorical inconsistencies between regions; In (b), the overall semantics of the news text conflict with specific phrases when placed in the realistic context. (c) mismatched text and images from unrelated real news sources, creating inconsistent narratives.}
    \label{fig:fake_examples}
\end{figure}

% current Limitations & challenges & research Objectives & research method

% limitations: 
% \textbf{} \textit{}
% 1. fusion different modalities to exploit the supplementary information, however it can not solve the matching but generated fake news (matching)； (a)(c) ---- 

% 2. external information: increased model complexity and training cost; dependent the accuracy of external knowledge.

% to solve these issues: intuitively, we think fake and real news exhibit different distributions or statictic charactoristics. therefore, we design MIANet to learn disrciminant representations. Specifically...

% module 1 matching, module 2 (a)(c).

% for mismatched fake: we aim to exploit the semantic inconsistency.
% for intra fake: we aim to exploit the discriminant features based on statistical charactoristics. 

% \textbf{} research challenge 1; evidence

% transition section

Research on fake news detection has evolved from relying exclusively on text-based approaches~\cite{yu2017convolutional,wu2024unified} to incorporating multimodal content, driven by the increasing prevalence of multimedia content on social media platforms~\cite {yu2022deep}. However, the lack of a systematic categorization of news content poses a substantial barrier to advancing effective multimodal fake news detection methods. To address this limitation, this study introduces a novel perspective that classifies multimodal fake news into two distinct categories: a) \textbf{fabricated fake news} encompasses cases where textual and visual content appear consistent but involve deliberate manipulations or synthetic generation in one or more modalities. It presents a more formidable detection challenge, particularly given the rapid advancements in generative AI technologies~\cite{zhao2023chatspot,li2023blip}, as demonstrated in Figure~\ref{fig:fake_examples} (a) and (b); b) \textbf{mismatched fake news} refers to instances where textual and visual content are semantically or logically inconsistent. These inconsistencies constitute a form of misinformation capable of distorting public perception and influencing behavior, as illustrated in Figure~\ref{fig:fake_examples} (c).

On the one hand, detecting fabricated fake news is highly challenging, yet research in this area remains limited. While some methods leverage external knowledge to improve content understanding, they often rely heavily on the timeliness and relevance of such knowledge, introducing potential biases that may undermine performance~\cite{wu2023see,zhangnatural}. Therefore, it derives the main research question of our study, \textit{\textbf{RQ 1: How to exploit the intrinsic discriminative information within each modality to identify fabricated fake news without relying on external knowledge?}}
On the other hand, despite recent advancements in detecting mismatched fake news, current methods still face notable limitations. These methods can be broadly categorized into auxiliary task-based approaches~\cite{zhou2020similarity,chen2022cross} and attention mechanism-based approaches~\cite{qian2021hierarchical,wei2022cross}. The former quantifies the similarity between modality representations in a shared subspace or fuses multimodal representations, while the latter captures cross-modal feature similarities to identify inconsistencies. However, both approaches often fail to extract explicit inconsistency or distinguish the inter-modal relationships into consistency and inconsistency features, leading to confusion in the model.
Thus, our second research question can be summarized as, \textit{\textbf{RQ 2: How to effectively detect inconsistencies in mismatched fake news while avoiding the overemphasis on cross-modal similarities and implicit pseudo-consistency?}} 

To address these research questions, we propose the Multimodal Inverse Attention Network (MIAN), which leverages unimodal and multimodal intrinsic discriminative information through hierarchical interaction and explicit inconsistency without relying on external knowledge. Specifically, to capture the intrinsic discriminant features of each modality, we build a hierarchical learning module consisting of two aspects: 1) Local-to-Local, which focuses on the contextual associations within the local features; and 2) Local-to-Global, which models the relationships between global and local features. This module fully explores unimodal information from news text and images and produces refined modality-specific representations to aid MIAN in detecting fabricated fake news, i.e., targeting for \textit{\textbf{RQ 1}}. To fuse multimodal features, we propose a cross-modal interaction module that facilitates interaction between the enhanced local features from different modalities, enabling effective detection of mismatched fake news, i.e., targeting for \textit{\textbf{RQ 2}}. Additionally, both the hierarchical learning module and the cross-modal interaction module equip our designed inverse attention mechanism, allowing MIAN to extract explicit inconsistent features that are crucial for fake news detection.
Overall, the contributions of this research can be summarized as:
\begin{itemize}
    % overall framework and application significance
    \item{We propose MIAN, a multimodal framework that leverages intrinsic discriminant information to improve news content-based fake news detection while offering a novel perspective for capturing discriminative relationships.}
    % two modules for intra and inter feature learning
    \item{We design a hierarchical learning module that enhances unimodal features through hierarchical relationships and a cross-modal interaction module that enables deep integration of inter-modal dependencies, improving both intra- and inter-feature learning.}
    % inverse attention and experiments
    \item{We present an inverse attention mechanism to detect inconsistencies across multiple levels, enabling the model to effectively identify diverse fake news. Extensive experiments demonstrate that MIAN outperforms state-of-the-art methods in accuracy.}
\end{itemize}

\begin{figure*}[!h]
	\centering
	\includegraphics[width=\textwidth]{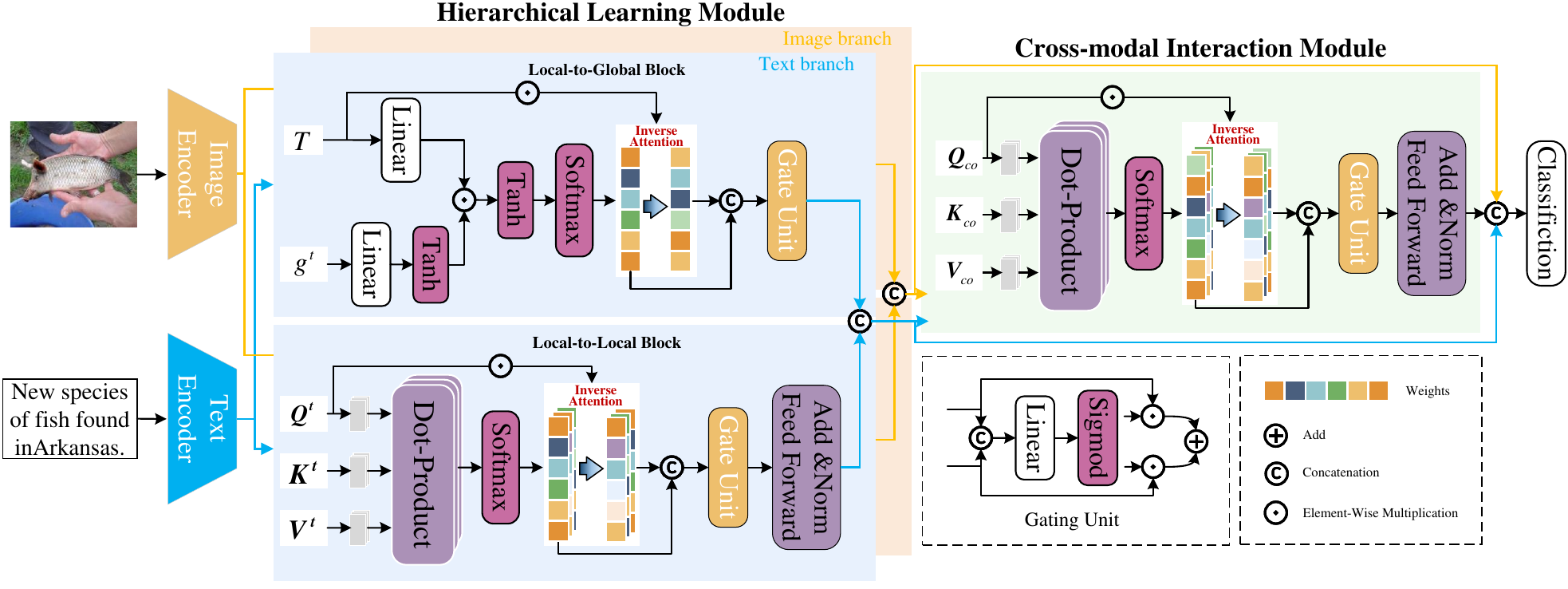}
	\caption{The proposed framework, MIAN, aims to detect fake news by fully leveraging both textual and visual content in news articles. Given a news piece, the model first utilizes modality-specific encoders to extract unimodal embeddings. Next, we apply a hierarchical learning module with different attention mechanisms in the Local-to-Global and Local-to-Local Blocks to capture and enhance hierarchical feature interactions. The enhanced unimodal features are then input into a co-attention mechanism to generate multimodal fused features. Throughout the various attention mechanisms, we incorporate an inverse attention module to explicitly extract inconsistencies between different targets. Finally, all enhanced unimodal and multimodal features are fused for fake news detection.}
	\label{fig:Overview}
\end{figure*}

\section{Related Work}
% This section focuses on multimodal fake news detection methods relevant to the proposed approach, while a detailed description of unimodal methods is provided in the appendix \ref{unimodality works}.

\subsection{News Content-based Methods}
The lack of correlation between textual and visual content in news is a key characteristic of certain types of multimodal fake news. Consequently, several studies have focused on measuring multimodal consistency to verify the credibility of news. Zhou et al. \cite{zhou2020similarity} utilized a pre-trained image captioning model to convert images into textual descriptions which achieved cross-modal semantic space alignment, then assessed the multimodal consistency between the original text and the generated captions. Similarly, Xue et al.~\cite{xue2021detecting} employed shared weights to enforce the alignment of textual and visual representations within a unified semantic space, calculating the similarity of the transformed multimodal representations. To mitigate the semantic gap between modalities, Jiang et al. \cite{jiang2023similarity} leveraged CLIP \cite{radford2021learning} to extract features from news text and images, subsequently using cosine similarity to guide multimodal fusion. Chen et al. \cite{chen2022cross} introduced cross-modal contrastive learning as an auxiliary task, and then used the Kullback-Leibler (KL) divergence to measure the ambiguity score between the latent distributions of text and image sampled from the autoencoder. Qi et al.~\cite{qi2021improving} measured multimodal entity inconsistency by calculating the similarity between entities in the news text and the corresponding visual content. 

Under limited available information, some methods refine unimodal features and employ hierarchical cross-attention mechanisms \cite{lu2019vilbert} to model inter-modal relationships. Qian et al. \cite{qian2021hierarchical} extracted text features at different hierarchical levels and fused them with visual features using a contextual transformer to learn cross-modal contextual features and supplementary information. Wu et al.~\cite{wei2022cross} extracted spatial and frequency domain features from images and progressively fused them with textual features through stacked cross-attention mechanisms.
However, their approaches primarily focus on exploring inter-modal relationships. Even though a few methods attempt to enhance unimodal feature representations, they overlook intra-modal interactions, which are crucial for detecting fake news, particularly fabricated news. Moreover, most existing methods model high-level semantic interactions through implicit pseudo-consistency, making it challenging to explicitly capture cross-modal inconsistencies.

\subsection{External Knowledge-based Methods}
Due to limitations in text length or the emergence of novel terms, discrepancies in understanding news content may arise. As a result, several studies have incorporated external knowledge extracted from social media \cite{zhang2021mining}, knowledge graphs \cite{zhang2019multi,dun2021kan} or internet retrieval \cite{zhang2024escnet,qi2024sniffer,yu2020multi} to enhance the performance of fake news detection. Zheng et al.~\cite{zheng2022mfan} employed an attention mechanism to combine social graph features, composed of user and comment data, with textual and visual features of news, generating discriminative features for fake news detection. Wu et al.~\cite{wu2023see} performed the deep fusion of text and image features based on three distinct reading patterns while leveraging the relationship between news comments and content to capture semantic inconsistencies. Zhang et al.~\cite{zhang2023hierarchical} leveraged news text entities as prompts to guide a large-scale vision-language model in generating image entities that enhance the semantic knowledge of the content. Zhang et al.~\cite{zhangnatural} converted news images and externally relevant knowledge into pure textual representations, which were then combined with the original text and fed into a prompt-based large language model to predict the authenticity of the news. While some methods enhance the representation of news content by incorporating external knowledge, this simultaneously introduces the risk of noise, which is crucial for the accuracy of fake news detection. These challenges significantly hinder the progress of multimodal fake news detection.

% \section{Problem Definition}

\section{Methodology}

Fake news detection based on multimodal content aims to leverage both textual and corresponding visual information to evaluate the authenticity of news. Given a dataset of fake news detection $ D = \{ X,Y\} $, where $X = [{X^T},{X^V}]$ consists of the textual component $X^T$ and the visual component $X^V$, and $Y \in \{ 0,1\}$ is the corresponding label indicating whether the news is fake (0) or real (1). The objective of the model is to learn a mapping function $F:({X^T},{X^V}) \to Y$ from the training set of $D$, which maps the input space $({X^T},{X^V})$ to the labels $Y$. The learned function $F$ is then used to predict labels for the test instances.

Our MIAN aims to fully learn intrinsic discriminant features within and across modalities. The detailed architecture is illustrated in Figure~\ref{fig:Overview}. We first extract unimodal embeddings from the news text and images using modality-specific encoders. For each multimodal news instance, the text is represented as a sequence of ${m}$ tokens ${\boldsymbol{T} = \{ {t_1},{t_2},...,{t_m}\} }$, where ${\boldsymbol{T} \in {\mathbb{R}^{m \times d}}}$ and each token ${{t_i} \in {\mathbb{R}^{d}}}$ is a ${d}$-dimensional vector obtained from a pre-trained BERT model~\cite{devlin2018bert}. Meanwhile, we employ a pre-trained ViT model~\cite{dosovitskiy2020image} to encode the image content into ${u}$ visual tokens $\boldsymbol{O} = \left\{ {{o_1},{o_2},...,{o_u}} \right\}$, where${\boldsymbol{O} \in {\mathbb{R}^{u \times d}}}$. We then introduce a hierarchical learning module consisting of two components, the Local-to-Global Block and the Local-to-Local Block. This module enables multi-granularity interactions to enhance unimodal representations for both text and image content. A detailed description is provided in Section~\ref{section:HLM}. Subsequently, MIAN fuses the enhanced multimodal representations using a co-attention mechanism in the cross-modal interaction module. A more detailed description is given in Section~\ref{section:CIM}. Additionally, to enable the model to explicitly learn inconsistency features, we design an inverse attention mechanism and integrate it into both the hierarchical learning module and the cross-modal interaction module. The specifics are discussed in Section~\ref{section:IA}. Finally, the enhanced modality-specific representations from the hierarchical learning modules in both the text and image branches, along with the fused representation from the cross-modal interaction module, are concatenated to form the final news representation, which is then fed into the classifier to assess the news authenticity.
    
\subsection{Hierarchical Learning Module (HLM)}
\label{section:HLM}
To comprehensively explore the internal relationships within news text and image representations and to produce enhanced unimodal features, we propose the Hierarchical Learning Module (HLM). After obtaining modality-specific embeddings from pre-trained encoders for text and images, HLM further models intra-modal associations at different levels from two perspectives: the Local-to-Local Block, which captures contextual interactions among local features, and the Local-to-Global Block, which establishes dependency relationships between local and global features.

\noindent
\textbf{Local-to-Local Block.}
We employ a multi-head attention mechanism \cite{vaswani2017attention} to learn consistent relationships between tokens and capture intrinsic structural features, thereby generating context-aware, refined local representations within each unimodal branch. Specifically, for the text modality, given the text embedding $\boldsymbol{T}$, we first perform different multi-head linear transformations to derive the query, key, and value matrices: ${\boldsymbol{Q}_t^i} = \boldsymbol{TW}_{ll - q}^i$, ${\boldsymbol{K}_t^i} = \boldsymbol{TW}_{ll - k}^i$ and ${\boldsymbol{V}_t^i} = \boldsymbol{TW}_{ll - v}^i$, and then concatenate the outputs from the attention mechanism.
\begin{equation}
    \label{c-ll-att}
    \boldsymbol{Att}_t^i = softmax\left( {\frac{{{\boldsymbol{Q}}_t^i{\boldsymbol{K}}{{_t^i}^ \top }}}{{\sqrt {{d_k}} }}} \right),
\end{equation}
\begin{equation}
    \label{ll-cat}
    \boldsymbol{h}_t^i = \boldsymbol{Att}_t^i{\boldsymbol{V}_t^i},
\end{equation}
\begin{equation}
    \label{mh}
    \boldsymbol{\bar R}_t = \left[ {h_t^1,...,h_t^n} \right]{\boldsymbol{W}_{cat}},
\end{equation}
where all $\boldsymbol{W}$ are learnable parameters, and $d_k$ denotes the dimensionality. The $n$ weighted value vectors are concatenated using $\left[  \cdot  \right]$. Finally, a series of residual connections and feedforward layers are applied to produce the context-aware, local-to-local enhanced features.
\begin{equation}
	\boldsymbol{\hat R}_t^{ll} = LN(\boldsymbol{T} + \boldsymbol{W}_t^{fc1}\boldsymbol{\bar R}_t),
\end{equation}
\begin{equation}
	\boldsymbol{R}_t^{ll} = LN\left( {\boldsymbol{\hat R}_t^{ll} + ReLU\left( {\boldsymbol{W}_t^{fc2}\boldsymbol{\hat R}_t^{ll} + b} \right)} \right),
\end{equation}
where $\boldsymbol{W}_{fc1}^t$, $\boldsymbol{W}_{fc2}^t$ and $b$ are learnable parameters. $LN(\cdot)$ and $ReLU(\cdot)$ refer to layer normalization and activation function.

Similarly, the ViT encoder transforms the segmented image into a sequential representation of local features, enabling the application of the self-attention mechanism. Following the same procedure as described above, we derive $\boldsymbol{R}_o^{ll}$.

\noindent
\textbf{Local-to-Global Block.}
This block refines local features by integrating global context, using global features to bridge fine-grained local information with overarching semantics. Specifically, for the visual modality, given the image embedding $O$, we construct the global feature to encapsulate the overall semantics of the image from two complementary perspectives. On the one hand, the mean-pooled vector ${g^{mean}} = {1 \mathord{\left/{\vphantom {1 n}} \right.\kern-\nulldelimiterspace} n}\sum\limits_{i = 1}^n {{o_i}}$ summarizes the spatial information of the entire image. On the other hand, we use the $\left[ {cls} \right]$ token ${g^{cls}}$ output from the modality-specific encoder to represent the semantic information of the salient objects in the image. Finally, these two representations are concatenated to form the global feature: 
\begin{equation}
    \label{equ: global cat}
    {g_o} = \left[ {{g^{mean}},{g^{cls}}} \right].
\end{equation}

In this block, the global features serve as the query guidance vector, while the local features function as the matched vectors. Their inner product is computed to measure the similarity between the global and local features. The resulting similarity scores are normalized via a softmax function, ensuring that the attention weights across all regions sum to 1. Specifically, the attention weights reflecting the consistency between the global high-level semantic features and the word-level semantic features of the text are computed as follows:
\begin{equation}
    \label{c-lg}
    h_o = tanh(\boldsymbol{W}_o^1{\boldsymbol{O}} \odot tanh(\boldsymbol{W}_o^2{g_o})),
\end{equation}
\begin{equation}
    \label{c-lg-att}
    att_o^{lg} = softmax \left( {\boldsymbol{W}_o^3h_o} \right),
\end{equation}
\begin{equation}
    \boldsymbol{R}_o^{lg} = att_o^{lg}{\boldsymbol{O}},
\end{equation}
where all $\boldsymbol{W}$ are learnable parameters, $h_o$ denotes the hidden state of the text attention function, and $ \odot $ represents the element-wise multiplication. The process of calculating the weighted relationship $\boldsymbol{R}_t^{\lg }$ between global and local information in the news text branch follows the same procedure as in the image branch.

Finally, we concatenate the enhanced modality-specific features produced by the Local-to-Local Block and Local-to-Global Block in each modality branch to obtain the hierarchically enhanced representations. Specifically, for the text modality, the representation is ${\boldsymbol{R}_t} = [ {\boldsymbol{R}_t^{ll},\boldsymbol{R}_t^{\lg }} ]$, and for the visual modality, it is ${\boldsymbol{R}_o} = [ {\boldsymbol{R}_o^{ll},\boldsymbol{R}_o^{\lg }}]$.

\subsection{Cross-modal Interaction Module (CIM)}
\label{section:CIM}
To fuse multimodal representations, we propose the Cross-modal Interaction Module (CIM), which leverages a co-attention mechanism to capture inter-modal dependencies and effectively integrate the hierarchically enhanced representations of news text and images. Specifically, to obtain the text features enriched with image information, given the hierarchical enhanced representations $\boldsymbol{R}_t$ and $\boldsymbol{R}_o$ from the HLM, we apply distinct multi-head linear projections
$\boldsymbol{Q}_t^{co} = \boldsymbol{R}_t\boldsymbol{W}_t^{co-q}$, 
$\boldsymbol{K}_t^{co} = \boldsymbol{R}_t\boldsymbol{W}_t^{co-k}$ 
and $\boldsymbol{V}_t^{co} = \boldsymbol{R}_t\boldsymbol{W}_t^{co-v}$ to derive the query, key, and value matrices, respectively. Due to the similarity of operations to the Local-to-Local Block and space constraints, the specific details of the multi-head attention mechanism are omitted here.
\begin{equation}
    \label{c-co-att}
    \boldsymbol{Att}_{t \to o}^{co} = softmax \left( {\frac{{\boldsymbol{Q}_t^{co}\boldsymbol{K}{{_t^{co}}^ \top }}}{{\sqrt {{d_k}} }}} \right),
\end{equation}
\begin{equation}
     \boldsymbol{\bar R}_{t \to o}^{co} = \boldsymbol{Att}_{t \to o}^{co}{\boldsymbol{V}_t^{co}},
\end{equation}
\begin{equation}
    \boldsymbol{\hat R}_t^{co} = LN({\boldsymbol{R}_t} +\boldsymbol{W} _{co}^{fc1}\boldsymbol{\bar R}_{t \to o}^{co}),
\end{equation}
\begin{equation}
    \boldsymbol{R}_t^{co} = LN\left( {\boldsymbol{\hat R}_t^{co} + ReLU\left( {\boldsymbol{W}_{co}^{fc2}\boldsymbol{\hat R}_t^{co} + b} \right)} \right).
\end{equation}

Similarly, after applying the above process, the fused features $\boldsymbol{R}_o^co$ are obtained, which capture the influence of the image on the text.
\subsection{Inverse Attention}
\label{section:IA}
Exploring intrinsic relationships between modalities remains a significant challenge in multimodal fake news detection, especially in cases involving mismatched textual and visual content. Although the widely adopted cross-attention mechanism has shown promise in modeling cross-modal interactions, it often fails to capture inconsistencies between cross-modal features effectively. 

Specifically, cross-attention calculates the similarity between query and key vectors to determine which key vectors each query vector should attend to, and then applies the resulting similarity weights to compute a weighted sum of value vectors. While this mechanism effectively emphasizes similar content across modalities, it may inadvertently encourage the network to learn implicit pseudo-consistencies, leading to confusion in fake news detection. To address this limitation, we propose the inverse attention mechanism, which amplifies dissimilar features to enhance the learning of explicit inconsistency signals. The detailed explanation of the inverse attention mechanism employed in CIM is as follows:
\begin{equation}
    \label{equ:ic-co-att}
    \boldsymbol{Att}_{t \to o}^{co-ic} = softmax \left( {{\boldsymbol{A}} - {\bf{\boldsymbol{Att}}}_{t \to o}^{co}} \right),
\end{equation}
\begin{equation}
    \label{equ:ic-co-att-1}
     \boldsymbol{\bar R}_{t \to o}^{co-ic} = \boldsymbol{Att}_{t \to o}^{co-ic}{\boldsymbol{V}_t^{co}},
\end{equation}
where, $\boldsymbol{A}$ is the introduced scalar matrix, which is used to subtract from the attention matrix 
$\boldsymbol{Att}_{t \to o}^{co}$ The matrix $\boldsymbol{Att}_{t \to o}^{co-ic}$ represents the inverse attention weights, where the values corresponding to inconsistent vectors are larger. $\boldsymbol{\bar R}_{t \to o}^{co-ic}$ denotes the desired explicit inconsistency features. Given the consistency feature $\boldsymbol{\bar R}_{t \to o}^{co}$
and the inconsistency features $\boldsymbol{\bar R}_{t \to o}^{co-ic}$, which provide distinct information for fake news detection, we employ a gating unit to combine them effectively:
\begin{equation}
    \label{eq:weight_gate}
    g = \sigma \left( {{\boldsymbol{W}_g}\left[ {\boldsymbol{\bar R}_{t \to o}^{co},\boldsymbol{\bar R}_{t \to o}^{co-ic}} \right] + {b_g}} \right),
\end{equation}
\begin{equation}
    \label{eq:combine}
    \boldsymbol{R}_{t \to o}^{co} = g \cdot \boldsymbol{\bar R}_{t \to o}^{co} + \left( {1 - g} \right) \cdot \boldsymbol{\bar R}_{t \to o}^{co-ic},
\end{equation}
where, $\boldsymbol{W}_g$ and $b_g$ represent learnable parameters, while $\left[  \cdot  \right]$ denotes the concatenation operation. The function $\sigma $ is the sigmoid function, and $g$ is the gating weight computed during the process.

Given that inconsistencies may also emerge within the content of fake news texts or images, the inverse attention mechanism is incorporated into both the Local-to-Global and Local-to-Local modules. Specifically, during the computation of $\boldsymbol{R}_t^{ll}$, $\boldsymbol{R}_o^{ll}$, $\boldsymbol{R}_t^{lg}$, and $\boldsymbol{R}_o^{lg}$, similar operations to those described in Equations \ref{equ:ic-co-att} and \ref{equ:ic-co-att-1} are applied. This design enables the model to capture inconsistent features at both intra- and inter-modal levels, thereby improving its effectiveness in detecting various types of fake news. 

Notably, since interactions are performed at the token level, the values of individual elements directly influence the computation of inconsistency weights, particularly when processing textual embeddings. To alleviate the impact of irrelevant or missing information, we introduce a positional mask. Furthermore, due to the potential ambiguity in expression arising from the sequence order of text and the positional information of objects in images, and since the attention mechanism lacks an inherent notion of sequence order, we also incorporate positional encoding. The specific formulation is as follows:
\begin{equation}
    P{E_{\left( {pos,2i} \right)}} = \sin \left( {\frac{{pos}}{{{{10000}^{\frac{{2i}}{d}}}}}} \right),
\end{equation}
\begin{equation}
    P{E_{\left( {pos,2i + 1} \right)}} = \cos \left( {\frac{{pos}}{{{{10000}^{\frac{{2i}}{d}}}}}} \right),
\end{equation}
where $pos$, $i$ and $d$ represent the position index, dimension index, and the dimension of the positional encoding, respectively.

\noindent
\textbf{Loss function.} Finally, we concatenate the enhanced modal-specific features, enriched with intra-modal inconsistency signals, and the multimodal fused features, capturing inter-modal inconsistencies, to construct the news representation $\boldsymbol{R}_n$. This representation is subsequently input into a multilayer perceptron-based classifier to predict confidence scores, $\hat y = MLP({\boldsymbol{R}_n})$. We define the loss function $L$ using cross-entropy as follows:
\begin{equation}
    \label{loss}
    L =  - y\log \left( {{{\hat y}}} \right) - \left( {1 - y} \right)\log \left( {{1-{\hat y}}} \right).
\end{equation}

\begin{table*}[t]
	\centering
    \small
	\begin{tabular}{ccccccccc}
		\hline
		\multirow{2}{*}{Datasets}    & \multirow{2}{*}{Methods}           & \multirow{2}{*}{Acc.} & \multicolumn{3}{c}{Fake   News} & \multicolumn{3}{c}{Real   News} \\
        \cmidrule(lr){4-6}  \cmidrule(lr){7-9}
		&                                   &                           & Pre.  & Rec.  & F1 & Pre.  & Rec.  & F1 \\ \hline
		\multirow{8}{*}{Weibo17}
		& CAFE \cite{chen2022cross}    & 0.840    & 0.855      & 0.830   & 0.842    & 0.825      & 0.851   & 0.837    \\
		& CMC \cite{wei2022cross}  & 0.908      & \underline{0.940}      & 0.869   & 0.899    & 0.876      & \underline{0.945}   & \underline{0.907}    \\
		& BMR \cite{ying2023bootstrapping} & \underline{0.918}     & 0.882      & \textbf{0.948}   & 0.914    & \textbf{0.942}      & 0.870   & 0.904    \\
        & MRHF \cite{wu2023see} & 0.907  & 0.939 & 0.869    & 0.903      & 0.879    & 0.931     & 0.904     \\  
        & MSACA \cite{wang2024fake} & 0.903  & 0.935 & 0.873    & 0.903      & 0.872    & 0.935     & 0.902     \\
        & RaCMC\cite{yu2025racmc}    & 0.915     & 0.910     & \underline{0.924}     & \underline{0.917}     & 0.921     & 0.906     & 0.914     \\
		& \textbf{MIAN}   & \textbf{0.936}     & \textbf{0.950}     & 0.920   & \textbf{0.935}    & \underline{0.923}     & \textbf{0.952}   & \textbf{0.937}   \\ \hline
		\multirow{5}{*}{Weibo21}    & EANN \cite{wang2018eann}        & 0.870                & 0.902      & 0.825   & 0.862    & 0.841      & 0.912   & 0.875    \\
		& SpotFake \cite{singhal2019spotfake} & 0.851                     & \textbf{0.953}   & 0.733   & 0.828    & 0.786      & \textbf{0.964}   & 0.866    \\
		& CAFE \cite{chen2022cross}        & 0.882                     & 0.857      & \underline{0.915}   & 0.885    & 0.907      & 0.844   & 0.876    \\
		& BMR \cite{ying2023bootstrapping}          & \underline{0.929}        & 0.908      & \textbf{0.947}   & \underline{0.927}    & \underline{0.946}      & 0.906   & \underline{0.925}    \\
		& \textbf{MIAN}                          & \textbf{0.938}       & \underline{0.924}  & \textbf{0.947}   & \textbf{0.936}    & \textbf{0.950}      & \underline{0.928}   & \textbf{0.939}    \\ \hline
		\multirow{8}{*}{GossipCop}   & EANN \cite{wang2018eann}        & 0.864                     & 0.702      & 0.518   & 0.594    & 0.887      & 0.956   & 0.920    \\
		& SpotFake \cite{singhal2019spotfake} & 0.858         & 0.732      & 0.372   & 0.494    & 0.866      & 0.962   & 0.914    \\
		& CAFE \cite{chen2022cross}        & 0.867         & 0.732      & 0.490   & 0.587    & 0.887      & 0.957   & 0.921    \\
		& CMC \cite{wei2022cross}          & 0.893         & \underline{0.826}      & \underline{0.657}   & \underline{0.692}    & \underline{0.920}      & 0.963   & 0.935   \\
		& BMR  \cite{ying2023bootstrapping}         & \underline{0.895}         & 0.752      & 0.639   & 0.691    & \underline{0.920}     & \underline{0.965}  & \underline{0.936}    \\
        & MSACA \cite{wang2024fake} & 0.887     & 0.816    & 0.538    & 0.646      & 0.897    & \textbf{0.971}     & 0.933     \\
        & RaCMC\cite{yu2025racmc} & 0.879      & 0.745    & 0.563     & 0.641     & 0.902     & 0.954     & 0.927     \\
		& \textbf{MIAN}   & \textbf{0.923}         & \textbf{0.834}      & \textbf{0.872}   & \textbf{0.853}    & \textbf{0.956}      & 0.941   &\textbf{0.948}   \\ \hline
		\multirow{6}{*}{PolitiFact} 
		& HMCAN  \cite{qian2021hierarchical}         & 0.864 	                & 0.738 	 & \textbf{0.933}   & 0.824 	  & \textbf{0.960} 	   & 0.828 	 & 0.889    \\ 
		& CAFE \cite{chen2022cross}          & 0.864         & 0.724      & 0.778   & 0.750    & 0.895      & 0.919   & 0.907    \\
		& LII  \cite{singhal2022leveraging}        & 0.907      & \underline{0.895}      & 0.872   & \underline{0.883}    & 0.900      & 0.918   & 0.909	\\	
		& CMC  \cite{wei2022cross}           & 0.893                     & 0.826      & 0.657   & 0.692    & 0.920      &\underline{0.963}   & 0.935    \\
        & RaCMC\cite{yu2025racmc} & \underline{0.922}      & 0.833    & \underline{0.926}     & 0.877     & \underline{0.967}     & 0.921     & \underline{0.943}     \\ 
		& \textbf{MIAN}                     & \textbf{0.953}        & \textbf{0.971}     & 0.919   & \textbf{0.944}    & 0.941      & \textbf{0.980}   & \textbf{0.960}    \\ \hline
	\end{tabular}
    \caption{Results of comparison among different approaches on Weibo17, Weibo21, GossipCop and PolitiFact Datasets. The best performance is highlighted in bold, and the follow-up is highlighted in underlined.}
    \label{tab:comparison}
\end{table*}

\section{Experiments}
The experiments in this study aim to address several key challenges in fake news detection by exploring the following research questions:
\begin{itemize}
    \item \textbf{Q1:} Can MIAN effectively improve the performance of fake news detection?
    \item \textbf{Q2:} Does each component contribute to improving detection?
    \item \textbf{Q3:} Can the proposed designs improve the accuracy of detecting specific types of fake news?
    \end{itemize}

\subsection{Configurations}
\textbf{Dataset.}
Our experiments are evaluated on four real-world datasets: Weibo17, Weibo21, GossipCop, and PolitiFact, which cover almost publicly available datasets. These datasets used in the experiments not only cover multiple domains but also include specific domain datasets where fake news frequently occurs. The detailed descriptions of these datasets are referred to Appendix \ref{appendix:datasets}.

\noindent
\textbf{Baselines.}
To evaluate the effectiveness and performance of the proposed method in multimodal fake news detection. The baseline methods include classic models that leverage deep neural networks for extraction such as EANN \cite{wang2018eann}, Spotfake \cite{singhal2019spotfake}, HMCAN \cite{qian2021hierarchical} and RaCMC \cite{yu2025racmc}, as well as methods exploring cross-modal relationships such as CAFE \cite{chen2022cross}, CMC \cite{wei2022cross}, BMR \cite{ying2023bootstrapping} and MSACA \cite{wang2024fake}, and integrating external knowledge for enhanced detection accuracy such as MRHF~\cite{wu2023see}. The detailed descriptions of these methods are referred to Appendix \ref{appendix:baselines}. 

\noindent
\textbf{Implementation.} 
All experiments were implemented using the PyTorch toolkit on an NVIDIA A100 GPU. We pad the sequence length of news text to 196 for all datasets, denoted as $m=196$, and put the processed text into the BERT. Images were resized to ${224 \times 224}$ pixels and subsequently divided into ${u = 14 \times 14}$ patches, which were then input into the vit-patch16-224. The multi-head attention used in both the Local-to-Local Block and the Cross-modal Interaction Module consists of 2 layers with 12 heads. The initial learning rate was set to $2{e^{ - 6}}$, and we utilized the StepLR decay strategy with a decay step of 20 and a decay rate of 0.5 to help the model converge better.

\subsection{Overall Performance}
Table \ref{tab:comparison} presents the performance results of the proposed method and the comparison methods on the public datasets. Although our method does not achieve the highest precision or recall, it outperforms the other comparison methods in terms of accuracy and F1 score, demonstrating its superior balance and overall performance, thus resolving \textbf{Q1}.

EANN and SpotFake, as early classic methods, simply concatenate the multimodal features obtained from modal-specific encoders to represent news, and do not perform well. This is likely due to their failure to further explore unimodal features and the lack of cross-modal interactions. CAFE and BMR highlight the importance of unimodal features in decision-making, with BMR achieving suboptimal results on certain datasets, suggesting that unimodal news content contributes to fake news detection. HMCAN and MSACA attempt to extract multi-level unimodal features from modal-specific encoders but do not yield strong performance, indicating that when exploring unimodal features, it is crucial to consider their specific role in the fake news detection task and fully leverage the interactions between them. Moreover, HMCAN exhibits a significantly higher F1 score for real news compared to fake news, which could be due to the use of a co-attention mechanism to model cross-modal information, where the model overemphasizes modal consistency and overlooks the stronger inconsistencies present in fake news.

\begin{figure}[!t]
    \centering
    \includegraphics[width=\columnwidth]{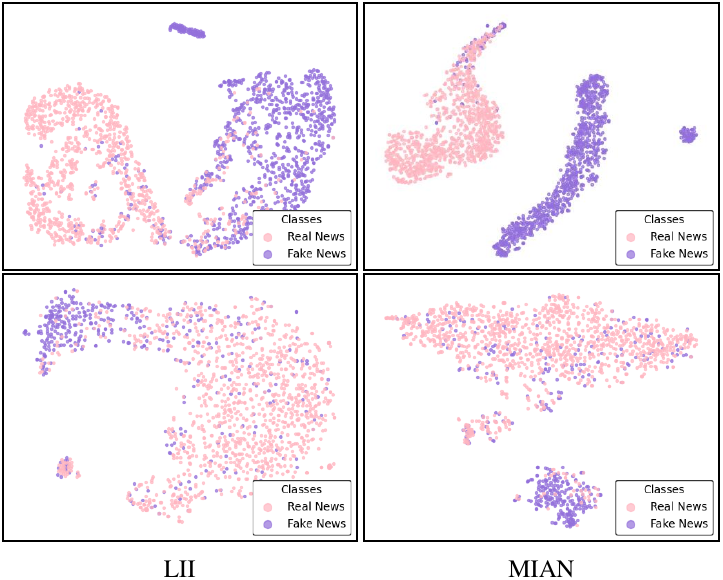}
    \caption{t-SNE visualization of the mined features on the test sets of Wei17 (first row) and GossipCop (second row).} %In the visualization, medium purple represents fake news features, while light pink denotes real news features.}
    \label{fig:t-SNE}
\end{figure}

\begin{table}[]
	\centering
    \small
	\begin{tabular}{ccccccccc}
		\hline
		\multirow{2}{*}{Datasets}  & \multirow{2}{*}{w/o models} & \multirow{2}{*}{Acc.}   & Fake   & Real &          \\
        % \cmidrule(lr){4}  \cmidrule(lr){5}
		&                          &                    & F1      & F1\\ \hline
		\multirow{5}{*}{Weibo17}  & \textbf{MIAN} & \textbf{0.936}   & \textbf{0.935}      & \textbf{0.937}  \\
		& w/o intra-lg             & 0.906    & 0.902         & 0.909    \\
		& w/o intra-lg-ic          & 0.910     & 0.907    & 0.912    \\
		& w/o intra-ll-ic          & 0.913     & 0.910   & 0.916    \\
		& w/o inter-ic             & 0.908     & 0.905   & 0.910     \\ \hline
	\end{tabular}
    \caption{The ablation experiment on the F1 score of the variant structure design on the Weibo17 dataset.}
    \label{tab:ablation}
\end{table}
Additionally, we provide the t-SNE visualization of the features for real and fake news in Figure \ref{fig:t-SNE}. Since the LII model also generates discriminative features through enhanced intra- and inter-modal methods, it serves as a comparison model to MIAN. It demonstrates that the features learned by the proposed model are more discriminative.

\subsection{Ablation Studies}

\noindent
\textbf{Impact of each component.} In order to address \textbf{Q2}, we evaluate MIAN against its variants by systematically removing specific components. Specifically, w/o intra-lg refers to eliminating the Local-to-Global Block, thereby removing the interaction between global and local features within both text and image branches. w/o intra-lg-ic and w/o intra-ll-ic refer to removing the inconsistent features generated by inverse attention at different levels within the same modality (Local-to-Global and Local-to-Local, respectively). w/o inter-ic refers to excluding the inconsistent features in the CIM.

Table \ref{tab:ablation} presents the results of the ablation experiments comparing various method variants, demonstrating the following key findings: (1) In the multimodal fake news detection task, comprehensive and in-depth unimodal information is essential. Extracting features from text or images aids the model in understanding the news content more effectively. (2) The loss of inconsistent information at any stage significantly impacts the accuracy of the detection results, particularly for fake news. Other metrics and results on the GossipCop dataset can be found in the Appendix \ref{appendix:experiments_full}.

\begin{table}[!t]
	\centering
    \small
	\begin{tabular}{ccccccccc}
		\hline
		\multirow{2}{*}{Methods}  & \multicolumn{1}{c}{Real}  & \multicolumn{2}{c}{Fake}  \\ 
		\cmidrule(lr){2-2} \cmidrule(lr){3-4} 
		& 0      & 4      & 5     & Avg.        &           \\ \hline
		HMCAN   & 0.776  & 0.804  & \underline{0.968} & 0.822           \\
		CAFE     & 0.907  & 0.794 & 0.938 & \underline{0.842}             \\
		LII    & \underline{0.909}  & \underline{0.840}   & 0.941 & 0.773               \\
		\textbf{MIAN}   & \textbf{0.920}  & \textbf{0.843}  & \textbf{0.972} & \textbf{0.870}     \\ \hline
	\end{tabular}
    \caption{Performance comparison to different models in terms of accuracy on the True, False Connection, and Manipulation Content classes of the Fakeddit dataset.}
    \label{tab:fakeddit_comparion}
\end{table}

\noindent
\textbf{Impact of different types.} To further validate the effectiveness of the proposed method in identifying both fabricated news and mismatched news, we introduce the Fakeddit dataset \cite{nakamura2019r}, which includes multiple fake news categories. In this dataset, label 0 represents real news, label 4 corresponds to False Connection, and label 5 denotes Manipulation Content, all of which align with the focus of our study. This experiment answers \textbf{Q3}, demonstrating that our specific design significantly enhances the accuracy of detecting various types of news.

For the comparative experiments, three representative methods were selected for comparison with our approach: HMCAN, which employs a co-attention mechanism to fuse multimodal information; CAFE, which utilizes the semantic differences between modalities as auxiliary tasks; and LII, which simultaneously leverages both intra-modal and inter-modal relationships. The results, presented in Table \ref{tab:fakeddit_comparion}, demonstrate that our method achieves superior performance. 

\begin{figure}[]
    \centering
    \includegraphics[width=0.98\columnwidth]{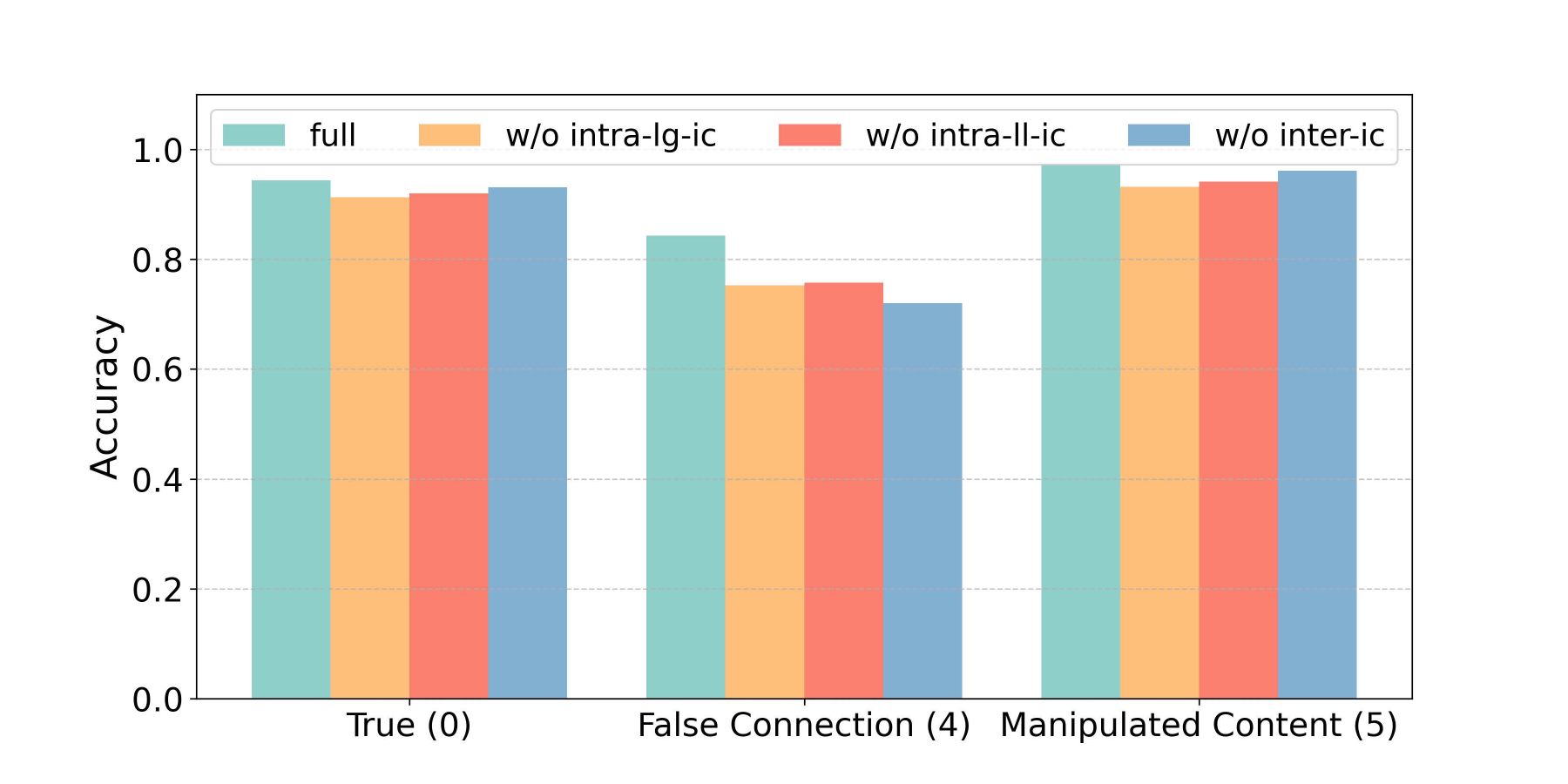}
    \caption{The performance of different variants of the proposed model in terms of accuracy on the True, False Connection, and Manipulation Content classes in the Fakeddit dataset.}
    \label{fig:fakeddit_ablation_3c}
\end{figure}

Furthermore, the ablation study, shown in Figure \ref{fig:fakeddit_ablation_3c}, reveals that removing unimodal modeling or inconsistency feature extraction results in notable accuracy degradation in categories 5 (Manipulated Content) and 4 (False Connection). This indicates that the proposed modules effectively identify both fabricated news and mismatched news. The details of the comparison and ablation experiments on other categories of the Fakeddit dataset can be found in Appendix \ref{appendix:fakeddit}.

\section{Conclusion \& Limitations}
In this paper, we propose MIAN, a novel multimodal fake news detection method that effectively identifies various types of fake news. MIAN explores hierarchical interactions within news text and image contents to generate enhanced unimodal representations, and then utilizes a co-attention mechanism to model inter-modal dependencies for multimodal feature fusion. Additionally, we introduce an inverse attention mechanism from a new perspective to explicitly learn and extract intra- and inter-modal inconsistencies. Experimental results demonstrate that the proposed method achieves optimal accuracy and effectively detects different types of fake news.

Despite the promising results, our method currently relies on a single image per news article, which could lead to information bias between the two modalities. Future work could address this by extending the approach to incorporate multiple images per news article. 

\section*{Acknowledgments}
This work was supported by the Joint Project for Innovation and Development of Shandong Natural Science Foundation, (No. ZR2022LZH012).

\clearpage
\newpage
\bibliographystyle{named}
\bibliography{ijcai25}

 \clearpage
\newpage
\appendix

\section*{Appendix}
\section{Datasets}
\label{appendix:datasets}
The dataset descriptions are as follows:

\begin{enumerate}
	\item{\emph{Weibo17:} The Weibo17 dataset consists of news posts collected from May 2012 to January 2016. The authenticity of the news is determined by the authoritative Chinese news agency Xinhua News Agency for real news, and by the official debunking system of Sina for fake news. Each post in the Weibo17 dataset includes three components: textual content, accompanying images, and social context. The proposed method in this study does not utilize external knowledge such as social context information, focusing solely on the textual and visual information of the news. Hence, all other components except the required two modalities are filtered out.}
	
	\item{\emph{Weibo21:} The Weibo21 dataset consists of news posts collected from December 2014 to March 2021, covering nine domains including Science, Military, Education, Disasters, Politics, Health, Finance, Entertainment, and Society. The fake news is publicly confirmed on the official platform of Weibo Community Management Center, while the real news is verified by NewsVerify8. Each post in the Weibo21 dataset includes news content, news images, user comments, publication time, and domain labels. Therefore, the processing procedure is similar to Weibo17.}
	
	\item{\emph{GossipCop:} The GossipCop is an entertainment news website dedicated to providing news reports and gossip information about celebrities, stars, movies, television, music, and other aspects of the entertainment industry. Additionally, the website verifies rumors and fake news in entertainment news and reveals the truth behind them. They verify or debunk various rumors and news reports in the entertainment industry through investigation, interviews with relevant individuals, and verification of reliable sources. The GossipCop dataset is collected from the website's publications from July 2000 to December 2018. The proposed method in this study applies deep learning techniques to analyze this dataset.}
	
	\item{\emph{PolitiFact:} The PolitiFact is a non-profit, independent news organization dedicated to fact-checking political statements and verifying their accuracy. They investigate and evaluate statements and claims made by politicians, government agencies, political organizations, and the media to reveal the truth and debunk misinformation. The PolitiFact dataset is collected from the website's publications from May 2002 to July 2018. The proposed method in this study leverages deep learning techniques for analyzing this dataset.}
\end{enumerate}

In the experiments, except for the Weibo17 dataset, which retains the same data splitting scheme as the baseline, the remaining datasets are divided into training and testing sets in a 9:1 ratio. Additionally, detailed statistical information on the preprocessed datasets is presented in Table \ref{tab:table1}.

\begin{table}[h]
	\centering
	\begin{tabular}{cccccc}
		\hline
		\multirow{2}{*}{Datasets} & \multicolumn{2}{c}{Train set}    & \multicolumn{2}{c}{Test set}    & \multirow{2}{*}{Total} \\ \cmidrule(lr){2-3} \cmidrule(lr){4-5}
		& \multicolumn{1}{c}{Fake} & Real  & \multicolumn{1}{c}{Fake} & Real &                        \\ \hline
		Weibo17                  & \multicolumn{1}{c}{4168} & 4301  & \multicolumn{1}{c}{464}  & 478  & 9411                   \\ 
		Weibo21                  & \multicolumn{1}{c}{4038} & 4176  & \multicolumn{1}{c}{450}  & 464  & 9128                   \\ 
		GossipCop                & \multicolumn{1}{c}{3743} & 10355 & \multicolumn{1}{c}{417}  & 1152 & 15667                  \\ 
		PolitiFact               & \multicolumn{1}{c}{337}  & 444   & \multicolumn{1}{c}{38}   & 51   & 870                    \\ \hline
	\end{tabular}
    \caption{The statistics of datasets}
    \label{tab:table1}
\end{table}

\section{Baselines}
\label{appendix:baselines}
\begin{enumerate}
	
    \item{EANN \cite{wang2018eann}: It proposes a model consisting of a features extractor, a fake news detector, and a time discriminator, which is capable of extracting time-invariant features. This method utilizes adversarial neural networks to remove event-specific features from the fused textual-image features, enabling the transferability to newly emerging fake news. By effectively extracting and leveraging time-invariant features, EANN demonstrates improved performance in detecting fake news across different time periods.}

    \item{Spotfake \cite{singhal2019spotfake}}: In this algorithm, the pre-trained language models, such as BERT, is applied to extract textual content, while VGG19 is set up to obtain image features.
	
	\item{HMCAN \cite{qian2021hierarchical}: It proposes a novel approach for modeling contextual information in a hierarchical multimodal setting. This method combines multi-level textual features extracted from 12 layers of BERT with visual features extracted from ResNet. By utilizing a co-attention mechanism, HMCAN hierarchically models the contextual information across different modalities.}
	
	\item{LII \cite{singhal2022leveraging}: It proposes a novel structure that effectively identifies and suppresses information from weaker modalities while extracting relevant information from stronger modalities for each sample. It utilizes a modality-wise relation learning module to learn feature representations within each modality and capture the relations between different modalities.}
	
	\item{CAFE \cite{chen2022cross}: It addresses the ambiguity problem in multimodal data by evaluating the cross-modal ambiguity learning using Kullback-Leibler divergence. This method dynamically controls the importance of cross-modal features and unimodal features by leveraging the learned ambiguity scores. By incorporating the ambiguity scores, CAFE effectively manages the trade-off between cross-modal and unimodal information, resulting in improved performance in handling the ambiguity in multimodal data.}
	
	\item{CMC \cite{wei2022cross}:  It proposes a cross-modal knowledge distillation method for improving the performance of multimodal fake news detection. It achieves information fusion and sharing across different modalities through knowledge transfer and modality alignment.}
	
	\item{BMR \cite{ying2023bootstrapping}: It introduces textual, image pattern, image semantics, and multimodal consistency information to make predictions from multiple perspectives. It dynamically weighs the importance of each perspective and achieves adaptive weighting and guided feature fusion through the Improved Multi-gate Mixture-of-Expert network.}
    \item{MSACA \cite{wang2024fake}: It propose the Multi-scale Semantic Alignment and Cross-modal Attention (MSACA) network, which addresses the challenge of aligning text and image modalities at multiple scales. The model constructs hierarchical multi-scale image representations, enhances semantic consistency between text and image embeddings, and employs a cross-modal attention mechanism to select the most discriminative features. }
    \item{MRHF \cite{wu2023see}: It introduces Multi-reading Habits Fusion Reasoning Networks (MRHFR) to address the limitations of shallow multimodal fusion and modality inconsistency in fake news detection. The model incorporates a cognition-aware fusion layer, inspired by three cognitive reading habits, to deeply integrate multimodal features, and a coherence constraint reasoning layer to measure and resolve semantic inconsistencies between modalities.}
    \item{RaMCM \cite{yu2025racmc}: It introduces a multiscale residual-aware compensation module to effectively fuse cross-modal features, and a multi-granularity constraints module that refines classification by amplifying differences between real and fake news at both the news and feature levels. This approach improves feature alignment for real news and separation for fake news, enhancing the overall detection performance.}
\end{enumerate}

\section{Supplementary Experiments}
\subsection{The detailed experiments}
\label{appendix:experiments_full}
To further validate the reliability of our results, we conducted statistical significance tests to assess whether the observed performance improvements of MIAN over baseline methods are consistent and non-random. As shown in Table \ref{tab:ttest-results}, the results confirm that the improvements are statistically significant, supporting the robustness of our conclusions and reinforcing the credibility of the comparative evaluations.
\begin{table}[h]
\centering
\resizebox{\linewidth}{!}{
\begin{tabular}{llll}
\toprule
\textbf{Dataset} & {p-value (acc)} & {p-value (F-F1)} & {p-value (R-F1)} \\
\midrule
Weibo17     & 0.0254\textsuperscript{*  }  & 0.0184\textsuperscript{*  }   & 0.0151\textsuperscript{*  }  \\
Weibo21     & 0.0453\textsuperscript{*  }   & 0.0619                       & 0.0279\textsuperscript{*  }  \\
GossipCop   & 0.0002\textsuperscript{***}   & 0.0001\textsuperscript{***}  & 0.0006\textsuperscript{***}  \\
PolitiFact  & 0.0042\textsuperscript{** }   & 0.0377\textsuperscript{*  }  & 0.0117\textsuperscript{*  }  \\
\bottomrule
\end{tabular}}
\caption{Significance level is marked with \textsuperscript{*}: significant  (p-value $<$ 0.05) , \textsuperscript{**}:  very significant (p-value$<$0.01), \textsuperscript{***}: highly significant (p-value$<$0.001). Results without a mark are not significant.}
\label{tab:ttest-results}
\end{table}

We conducted ablation experiments on the Weibo17 and GossipCop datasets, and the results are shown in Table \ref{tab:ablation_full}. In the ablation experiments on both datasets, our model demonstrated strong performance on Weibo17 and GossipCop, particularly achieving optimal results in accuracy (Acc) and F1 score. Although on GossipCop, precision and recall did not surpass some of the baseline models, the optimal F1 score indicates that our method strikes a good balance between precision and recall. We attribute this observation to the nature of the GossipCop dataset, where mismatches between text and images are more prevalent in the entertainment domain. Future work will focus on further optimizing recall and precision, especially in cases of class imbalance or high data noise, to improve the model's generalization ability.

\begin{table*}[!h]
	\centering
	\begin{tabular}{ccccccccc}
		\hline
		\multirow{2}{*}{Datasets}  & \multirow{2}{*}{w/o models} & \multirow{2}{*}{Acc.} &           & Fake News &          &           & Real News &          \\
        \cmidrule(lr){4-6}  \cmidrule(lr){7-9}
		&                          &                           & Pre. & Rec.    & F1  & Pre. & Rec.    & F1\\ \hline
		\multirow{5}{*}{Weibo17}  & \textbf{MIAN}                 & \textbf{0.936}                     & \textbf{0.950}      & \textbf{0.920}      & \textbf{0.935}    & \textbf{0.923}     & \textbf{0.952}     & \textbf{0.937}    \\
		& w/o intra-lg             & 0.906                     & 0.921     & 0.884     & 0.902    & 0.891     & 0.927     & 0.909    \\
		& w/o intra-lg-ic          & 0.910                      & 0.920      & 0.894     & 0.907    & 0.900       & 0.925     & 0.912    \\
		& w/o intra-ll-ic          & 0.913                     & 0.924     & 0.897     & 0.910     & 0.902     & 0.929     & 0.916    \\
		& w/o inter-ic             & 0.908                     & 0.918     & 0.892     & 0.905    & 0.898     & 0.923     & 0.910     \\ \hline
		\multirow{5}{*}{GossipCop} & \textbf{MIAN}                 & \textbf{0.901}                     & 0.872     & \textbf{0.755}    & \textbf{0.809}    & 0.909     & \textbf{0.961}     & \textbf{0.934}    \\
		& w/o intra-lg             & 0.895                     & 0.860      & 0.722     & 0.785    & 0.905     & 0.957     & 0.930     \\
		& w/o intra-lg-ic          & \textbf{0.901}                     & 0.856     & 0.734     & 0.780    & \textbf{0.915}     & 0.952     & 0.933    \\
		& w/o intra-ll-ic          & 0.893                     & 0.827     & \textbf{0.755}     & 0.789    & 0.914    & 0.943     & 0.928    \\
		& w/o inter-ic             & 0.898                     & \textbf{0.875}     & 0.719     & 0.789    & 0.905     & 0.960      & 0.932    \\ \hline
	\end{tabular}
    \caption{Ablation study on the architecture design of variant on Weibo17 and GossipCop datasets.}
    \label{tab:ablation_full}
\end{table*}

\subsection{Experiments on Fakeddit}
\label{appendix:fakeddit}
To further validate the effectiveness of our method, we introduced the Fakeddit dataset. Unlike the common datasets mentioned earlier, where each news article is only labeled as either real or fake, the Fakeddit dataset not only has labels indicating the authenticity of the news but also includes specific labels for five different types of fake news. These labels are True, Satire/Parody, Misleading Content, Imposter Content, False Connection, and Manipulate Content. Here is a brief explanation of each label:
\begin{itemize}
	\item{0 (True): True content is accurate in accordance with fact. Eight of the subreddits fall into this category, such as usnews and mildlyinteresting. The former consists of posts from various news sites. The latter encompasses real photos with accurate captions.}
	\item{1 (Satire/Parody): This category consists of content that spins true contemporary content with a satirical tone or information that makes it false. One of the four subreddits that make up this label is theonion, with headlines such as “Man Lowers Carbon Footprint By Bringing Reusable Bags Every Time He Buys Gas”.}
	\item{2 (Misleading Content): This category consists of information that is intentionally manipulated to fool the audience. Our dataset contains three subreddits in this category.}
	\item{3 (Imposter Content): This category contains two subreddits, which contain bot-generated content and are trained on a large number of other subreddits.}
	\item{4 (False Connection): Submission images in this category do not accurately support their text descriptions. We have four subreddits with this label, containing posts of images with captions that do not relate to the true meaning of the image.}
	\item{5 (Manipulated Content): Content that has been purposely manipulated through manual photo editing or other forms of alteration. The photoshopbattle subreddit comments (not submissions) make up the entirety of this category. Samples contain doctored derivatives of images from the submissions.}
\end{itemize}

\begin{figure*}[!h]
	\centering
	\includegraphics[width=\textwidth]{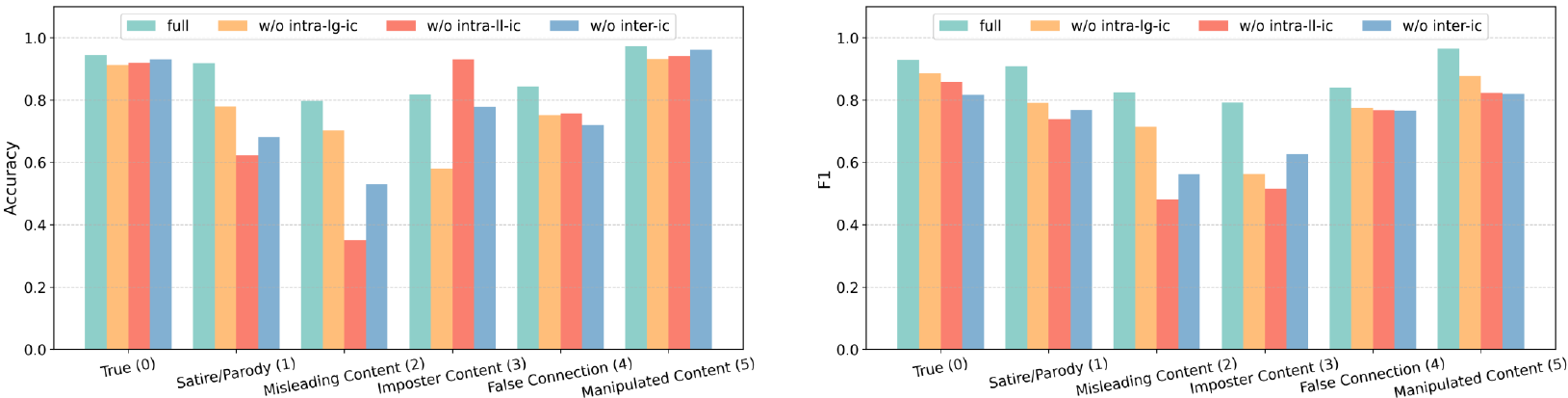}
	\caption{The performance of MIAN and its variants on the Fakeddit multiclass classification task.}
	\label{fig:fakeddit_ablation_6c}
\end{figure*}

Table \ref{tab:comparision_fakeddit_full} and Figure \ref{fig:fakeddit_ablation_6c} present the comparative and ablation experiment results of our approach across all categories in the Fakeddit dataset. In the comparative experiments, the proposed method achieved the best performance in the categories of True, Satire/Parody, Misleading Content, False Connection, and Manipulated Content, and second-best in the Imposter Content category. This demonstrates that MIAN has strong generalization ability across different types of fake news. In the ablation experiments, other model variants exhibited varying degrees of accuracy degradation, indicating that our proposed fake news classification approach effectively covers various fake news categories, and the introduced modules contribute to improving the model's performance.

\begin{figure*}[!h]
	\centering
	\includegraphics[width=\textwidth]{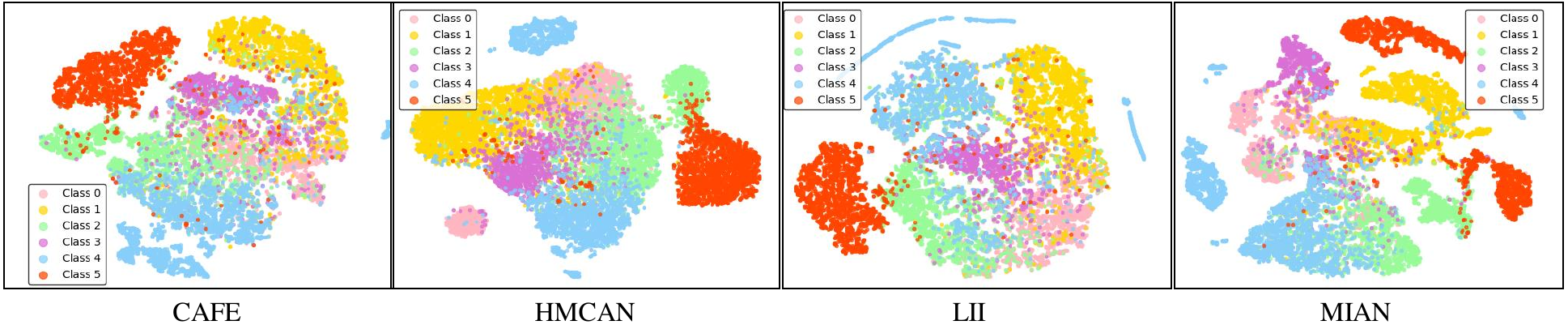}
	\caption{t-SNE visualization of the features extracted by the proposed method and other approaches on the Fakeddit dataset.}
	\label{fig:tSNE_fakeddit_comparison_6c}
\end{figure*}

The t-SNE visualizations of the features extracted by MIAN and other methods (CAFE, HMCAN, and LII) on the Fakeddit dataset, as shown in Figure \ref{fig:tSNE_fakeddit_comparison_6c}, highlight the superior performance of MIAN in separating the six classes of news. Unlike the other methods, MIAN demonstrates clear and distinct clustering, with minimal overlap between real and fake news categories. This indicates that MIAN effectively captures discriminative features across modalities, leading to improved classification accuracy. In contrast, CAFE, HMCAN, and LII exhibit more significant overlap, especially in distinguishing fake news types, suggesting that MIAN’s approach to hierarchical learning and co-attention is more effective in handling the complexities of multimodal fake news detection.
\begin{table*}[h]
	\centering
	\begin{tabular}{ccccccccc}
		\hline
		\multirow{2}{*}{Methods}  & \multicolumn{1}{c}{Real News}  & \multicolumn{6}{c}{Fake News}   & \multirow{2}{*}{Macro Avg.} \\ 
		\cmidrule(lr){2-2} \cmidrule(lr){3-8} 
		& 0      & 1      & 2      & 3      & 4      & 5      &  Micro Avg.      &           \\ \hline
		HMCAN    & 0.776  & 0.777 & 0.704  & 0.627 & 0.804  & \underline{0.968} & 0.822                                                 & 0.785                     \\
		CAFE     & 0.907 & \underline{0.866} & \underline{0.795} & \textbf{0.819} & 0.794 & 0.938 & \underline{0.842 }                            & \underline{0.853 }                   \\
		LII  & \underline{0.909} & 0.752 & 0.645  & 0.687 & \underline{0.840 }  & 0.941 & 0.773                                              & 0.802                     \\
		\textbf{MIAN}     & \textbf{0.920} & \textbf{0.918}  & \textbf{0.797} & \underline{0.818}  & \textbf{0.843}  & \textbf{0.972} & \textbf{0.870}                        & \textbf{0.881}               \\ \hline
	\end{tabular}
    \caption{Comparison of the accuracy of different models on Fakeddit dataset.}
    \label{tab:comparision_fakeddit_full}
\end{table*}

\end{document}